\documentclass[table]{article}
\usepackage{spconf,amsmath,graphicx}
\usepackage[dvipsnames]{xcolor}
\usepackage{comment}
\usepackage{hyperref}
\usepackage{url}
\usepackage{enumitem}
\usepackage{booktabs}
\usepackage{mathtools}
\usepackage{dirtytalk}
\usepackage{multirow}
\usepackage{soul}
\usepackage[font=small]{caption}

\title{E2E Segmentation in a Two-Pass Cascaded Encoder ASR Model}
\name{\begin{tabular}{c}W. Ronny Huang, Shuo-Yiin Chang, Tara N. Sainath, Yanzhang He, David Rybach, \\ Robert David, Rohit Prabhavalkar, Cyril Allauzen, Cal Peyser, Trevor D. Strohman\end{tabular}}
\address{Google Research}

\begin{document}
\ninept
\maketitle
\begin{abstract}
We explore unifying a neural segmenter with two-pass cascaded encoder ASR into a single model.
A key challenge is allowing the segmenter (which runs in real-time, synchronously with the decoder)
to finalize the 2nd pass (which runs 900 ms behind real-time)
without introducing user-perceived latency or deletion errors during inference.
We propose a design where the neural segmenter is integrated with the causal 1st pass decoder to emit a end-of-segment (EOS) signal in real-time.
The EOS signal is then used to finalize the non-causal 2nd pass.
We experiment with different ways to finalize the 2nd pass,
and find that a novel dummy frame injection strategy allows for simultaneous high quality 2nd pass results and low finalization latency.
On a real-world long-form captioning task (YouTube),
we achieve 2.4\% relative WER and 140 ms EOS latency gains over a baseline VAD-based segmenter with the same cascaded encoder.
\end{abstract}

\vspace{-6pt}
\section{Introduction}
\label{sec:intro}

Two-pass cascaded encoder models \cite{narayanan2021cascaded} are used in streaming ASR systems to provide real-time, causal results to the user from the 1st pass,
while also providing superior, high-quality, \textit{non-causal} results from a 2nd pass that lags the 1st pass by, e.g., 900 ms in order to access 900 ms of right-context frames.
Running in parallel allows the 2nd pass to, in particular, operate on long-form utterances (i.e., those longer than a few minutes) prevalent in captioning and dictation tasks.

Another common practice when decoding long-form utterances is to segment the utterance into shorter lengths,
and to deliver the final 2nd pass result at each end-of-segment (EOS).
Many downstream tasks (e.g. machine translation) can incrementally consume partial recognition results,
so quick and frequent segmentation is beneficial for overall user experience \cite{shangguan2021dissecting}.
In streaming settings, the segmenter runs synchronously with beam search and decides where to end each segment in real time.
In particular, it emits a EOS token, which resets the beam search, discarding all hypotheses except for the top hypothesis, whose state is passed to the next segment.
This clears room for new hypotheses to enter the beam, introducing more diversity.
More critically, the EOS token \textit{finalizes} the top hypothesis by discarding all other possible contenders, guaranteeing that the top hypothesis will no longer change,
e.g., by later evidence causing a non-top hypothesis to be promoted to the top.
Finalization is prerequisite for delivering the 2nd pass result to downstream tasks because downstream tasks typically require the hypothesis to be fixed.
On the other hand, a poorly placed EOS can lead to the wrong hypothesis being finalized, hurting ASR quality.
Consequently, the segmenter's job of deciding \textit{when} to emit EOS is foundationally important to both WER and user-perceived latency in long-form tasks.

Conventionally, EOS is determined by the VAD, a classifier predicting whether each frame is speech or silence \cite{ramirez2007voice,yoshimura2020end}.
A hand-coded state machine keeps track of the predictions, does some smoothing, and emits an EOS signal when, e.g. 200 ms in our case, of consecutive silence frames are detected.
There are two disadvantages to the VAD segmenter:
(1) it requires a state machine with hand-crafted hyperparameters which adds complexity, 
and (2) the VAD conditions on acoustic features only (speech/silence).
The E2E segmenter \cite{huang2022e2e} tackles these issues by predicting the EOS signal directly from the decoder of a streaming E2E ASR model.
Specifically it adds an additional RNNT joint layer trained to predict the EOS token via a weakly supervised EOS annotation method \cite{chang2022turn}.
This simplifies the system by unifying ASR and segmentation into a single model.
It also improves quality and EOS latency by leveraging both acoustic and linguistic context available to the E2E ASR model.
That said, E2E segmentation has only been demonstrated in single-pass models \cite{huang2022e2e}.

\begin{figure*}
  \centering
  \vspace{-31pt}
  \includegraphics[width=.326\linewidth]{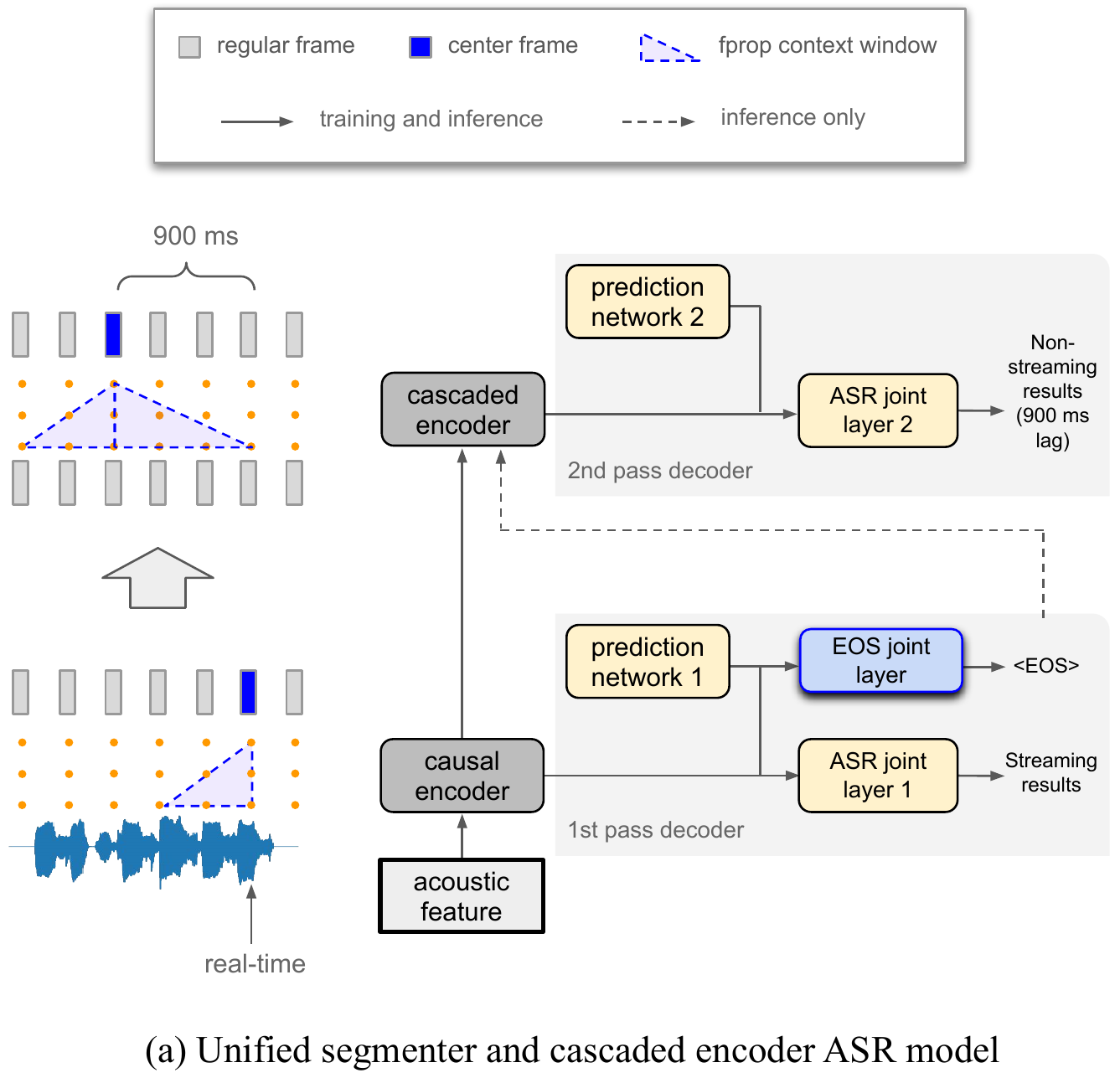}
  \quad
  \includegraphics[width=.644\linewidth]{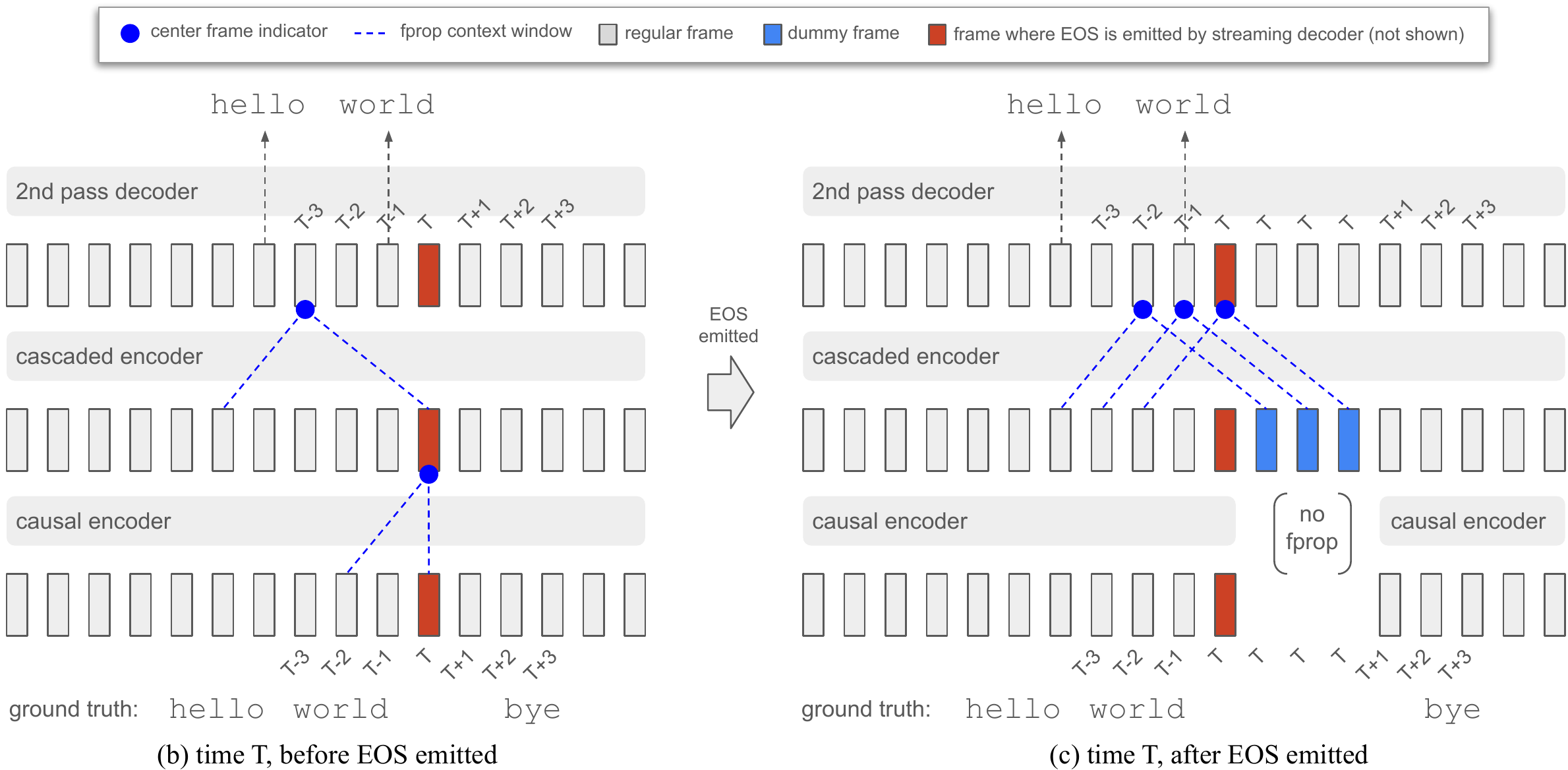}
  \vspace{-2pt}
  \caption{Schematic of model architecture (a) and inference procedure immediately before (b) and after EOS emission (c).}
  \vspace{-12pt}
  \label{fig:schem}
\end{figure*}

Unifying E2E segmentation with the higher-quality two-pass cascaded encoder model is the next logical step
toward building a low-latency and state-of-the-art streaming ASR system for long-form tasks.
In this work, we discuss the challenges to unifying ASR and segmentation and present our solution which led to a single cascaded-encoder model that predicts EOS as shown in Figure \ref{fig:schem}a,
with gains of 2.4\% WER relative, 7.3\% oracle WER relative, and 140 ms EOS latency on a YouTube long-form testset.

\vspace{-6pt}
\subsection{Related work}
\vspace{-4pt}
Previous works have sought to tune the 2nd-pass latency in two-pass ASR.
\cite{zhang2020unified} modifies the conformer layers to have dynamic right context length and combines it with 2nd-pass rescoring to allow for tunable finalization latency in a two-pass model.
Similarly, \cite{mahadeokar2022streaming} leverages two non-causal encoders with variable input context sizes to allow for fast and slow outputs from the 2nd pass.

Other works have sought to unify additional tasks with two-pass ASR into a single model.
E2E prefetching \cite{chang2020prefetch} lowers fulfillment latency by querying the downstream response based on 1st-pass results and delivering that response if the 1st pass matches the eventual 2nd-pass finalized result.
E2E endpointing \cite{li2021better,bijwadia2022unified} lowers microphone closing latency by predicting end-of-utterance with the 1st pass and using that signal to finalize the 2nd pass.
Unifying prefetching and endpointing with two-pass ASR is important for applications where the utterances are short and require quick 2nd-pass results, such as the voice search. 
In a similar spirit, we for the first time unify \textit{segmenting} with two-pass ASR
for applications where the utterances are long and yet still require quick 2nd-pass finalization,
such as live captioning, dictation, or conversation agents.

\vspace{-4pt}
\section{Unifying E2E Segmentation with Cascaded Encoder}
\vspace{-4pt}

In contrast to conventional two-pass models where the 2nd pass must wait till the end-of-segment before decoding \cite{sainath2019two},
the cascaded encoder's 2nd pass decodes in parallel with the 1st pass,
but in a time-staggered fashion so that it can access right-context audio frames.
Specifically, the 2nd pass decoder lags real-time by 900 ms so that the cascaded encoder can condition on 900 ms worth of frames to the right of its center frame \cite{sainath2022improving,ding2022unified},
as illustrated in Figure \ref{fig:schem}a.
While E2E segmentation has been successfully demonstrated for single-pass models,
certain challenges stand in the way of integrating it into the two-pass cascaded encoder setup.
We now discuss these challenges and our solutions.
Note that we'll refer to ``frame'' as the representation at a particular time,
but the representation could correspond to the input, causal encoder output, or cascaded encoder output,
depending on the discussion context.

\vspace{-6pt}
\subsection{1st-pass EOS prediction, 2nd-pass finalization}
The EOS signal should be determined in real-time since any delay directly affects user-perceived latency.
This is easy to implement with a VAD segmenter, since the VAD is run upstream of the decoder.
In a E2E segmenter, however, the EOS is determined by the decoder itself.
If EOS was determined by the 2nd pass decoder, its timestamp would always be at least 900 ms behind real-time,
which is a prohibitive amount of latency.
We propose the following inference strategy:
\begin{enumerate}[leftmargin=*]
\item The 1st pass predicts EOS in real-time
\item When the EOS signal is emitted, finalize the 1st pass and route the EOS timestamp to the 2nd pass
\item When the 2nd pass receives the EOS timestamp, inject dummy frames and finalize (discussed further in Section \ref{sec:dummy})
\end{enumerate}
Figure \ref{fig:schem}a illustrates this architecture.

\vspace{-4pt}
\subsection{2nd-pass finalization strategy}
\label{sec:dummy}
Note importantly that the procedures discussed in the following section applies to \textit{inference} only, where latency is critical.
The \textit{training} the procedure is discussed in Section \ref{sec:eosjoint}.
Now consider as illustration Figure \ref{fig:schem}b.
When the time is $T$, ``hello world'' has been spoken.
At this time, the causal 1st pass decoder (not shown) decides to emit an EOS (red frame at time $T$) and it
immediately routes the EOS's timestamp, i.e. $T$, to the 2nd pass.
At that time, the cascaded encoder's center frame is at $T-3$, or 3 frames (900 ms in reality) behind the real-time frame at $T$,
so the 2nd pass decoder has yet to decode the word, ``\texttt{world}''.
When the 2nd pass receives the EOS timestamp, there are several possible responses:

\begin{table}[b]
\vspace{-10pt}
\caption{Comparison of 2nd-pass EOS handling strategies}
\vspace{-8pt}
\label{tab:dummy}
\centering
\resizebox{.99\columnwidth}{!}{%
\begin{tabular}{l|c|cc}
\toprule
                                    &                          & \multicolumn{2}{c}{2nd pass final. latency}                                \\
2nd pass EOS handling               & WER                      & Algorithmic & Computational                                                \\
\midrule          
B1: Finalize at EOS emission        & 17.99                    & 0           & $\sim$ 0                                                     \\
B2: Finalize at EOS timestamp       & 17.31                    & 900         & $\ge$900                                                       \\
E1: B2 + dummy frames (zeros)       & 18.65                    & 0           & 208 \cite{sainath2022improving}                              \\ 
E2: B2 + dummy frames (last)        & \textbf{15.86}           & \textbf{0}  & \textbf{208} \cite{sainath2022improving}                     \\ 
\bottomrule
\end{tabular}
}
\end{table}

\begin{enumerate}[leftmargin=*]
\item \textbf{Finalize immediately}.
This would result in the last 3 frames of the segment not being decoded,
resulting in the erroneous deletion of ``\texttt{world}'',
since the 2nd pass center frame lags that of the 1st pass by 3 frames and is still processing $T-3$ when 1st pass emits EOS.
See Figure \ref{fig:schem}b.
\item \textbf{Finalize when 2nd pass reaches EOS timestamp}.
This would mean waiting until the center frame of the 2nd pass decoder (currently at $T-3$) to reach $T$,
or equivalently, for the center frame of the 1st pass decoder or causal encoder (currently at $T$)
reaches $T+3$.
This can be visualized in Figure \ref{fig:schem}b as sliding the blue fprop context windows 3 frames rightward.
By then, ``\texttt{world}'' is correctly decoded by the 2nd pass decoder,
avoiding the deletion error from Option 1 above.
However, this requires waiting for 3 additional frames to be generated
because they are needed by the cascaded encoder as right-context.
Incurring 3 frames of latency (900 ms in reality) can be prohibitive for user experience.
\item \textbf{Inject 900 ms of dummy right-context frames, then finalize when 2nd pass reaches EOS timestamp}.
In this best-of-both-worlds alternative, we still allow the 2nd pass to decode until the EOS timestamp to avoid deletion errors,
but we prevent \textit{waiting} for 3 right-context frames by immediately injecting 3 \textit{dummy} right-context frames,
as illustrated in Figure \ref{fig:schem}c.
Since the dummy frames are injected all at once, the cascaded encoder can fprop them all in parallel,
as illustrated by the three fprop context windows in Figure \ref{fig:schem}c.
And they are injected at the input to the cascaded encoder, so we save some compute by not fpropping anything through the causal encoder.
All this can occur at time $T$ before the arrival of the next frame;
thus the \textit{algorithmic} latency of the process is zero.
The \textit{computational} latency (time for the cascaded encoder to fprop and decode the dummy frames), on the other hand,
is a relatively low 208 ms based on measurements with a Pixel 6 phone on a similar architecture \cite{sainath2022improving}.
Unlike Option 2, which has an algorithmically lower-bounded latency of 900 ms,
208 ms is an acceptable amount of user-perceived latency and can be improved with better hardware.

We experiment with two types of dummy frames:
\vspace{-4pt}
\begin{itemize}[leftmargin=*]
    \item \textbf{Zero}: all dummy frames are zero vectors
    \item \textbf{Last-frame}: all dummy frames are identical to last frame outputted by the causal encoder. The rationale is that the last frame occurs during the brief silence after the user speaks and is a better encoder representation of silence than zeros.
\end{itemize}
\end{enumerate}

Table \ref{tab:dummy} summarizes the WER and algorithmic and computational latency of 2nd pass finalization for all the strategies discussed in this section.
We pick as our final strategy the one which gives the best WER and finalization latency, namely dummy last-frame injection (E2).
Again, we apply this strategy only during inference.
\vspace{-5pt}
\section{Experimental Settings}

\vspace{-3pt}
\subsection{Model and dataset}
Our base cascaded encoder model is identical to that of \cite{ding2022unified} in vocabulary, architecture, optimizer setup, training data, codebase, and training hardware.
Briefly, the causal encoder, cascaded encoder, and separate decoders have 47M, 60M, and 4.4M parameters, respectively.
It's trained on 400k hours of English anonymized audio-text pairs from multiple domains, such as YouTube and anonymized voice search traffic.
Our data handling abides by Google AI Principles \cite{googleai}.

\vspace{-5pt}
\subsection{EOS joint layer}
\label{sec:eosjoint}

Our EOS joint layer is identical in size to, and initialized from the weights of, the fully trained ASR joint layer.
It consumes the outputs of the encoder and prediction network, allowing it to see both acoustic and linguistic context, as shown in Figure \ref{fig:schem}a.
It is fine-tuned on the same dataset but with the target transcripts augmented by EOS tokens.
The EOS annotation scheme is identical to that of \cite{huang2022e2e,chang2022turn}.
Briefly, EOS tokens are inserted at the end of the transcript for short-query utterances, such as voice search.
They're also inserted between two words for other utterances, such as YouTube,
when there is a long silence preceded by no hesitation or an even longer silence preceded by hesitation.
During inference, if the EOS token negative log-probability cost goes below a certain threshold,
the 1st-pass beam search will immediately finalize and the EOS timestamp will be routed to the 2nd pass for 2nd-pass finalization.

\vspace{-5pt}
\subsection{Beam search}
We use a frame-synchronous beam search for both passes.
The 1st and 2nd passes have beam sizes of 4 and 8, respectively, with pruning threshold of 5 for both.
At each frame, we apply breadth-first search for possible expansions similar to \cite{tripathi2019monotonic}, ignoring any expansion with a negative log posterior of 5 or greater, and limiting the search depth to 10 expansions. 

\vspace{-5pt}
\subsection{Voice activity detector}
Our pipeline contains a lightweight voice activity detector \cite{zazo2016feature} upstream of the E2E model.
In addition to being used as a segmenter in our VAD segmenter experiments,
it also functions as a frame filterer;
consecutive frames classified as silence are filtered out after the initial 200 ms of silence.
Frame filtering is turned on for all experiments to match edge deployment conditions.

\vspace{-5pt}
\subsection{Evaluation}
\label{sec:evaluation}
YouTube videos are  multi-domain (commentary, lectures, meetings, etc.) and are often long-form \cite{narayanan2019recognizing},
making captioning of YouTube audio an ideal task for our long-form study.
We evaluate on YT\_LONG, a standard YouTube testset used in \cite{Soltau2017,chiu2019comparison,chiu2021rnn} whose videos are sampled from YouTube video-on-demand.
YT\_LONG has 77 total utterances made of 22.2 hours and 207191 words, with median utterance length of 14.8 minutes.

\vspace{-10pt}
\section{Results}
\vspace{-5pt}
In this section, we report the 50th and 90th percentile segment lengths in seconds (SL50/90) and EOS latency in milliseconds (EOS50/90), along with 1st and 2nd pass WER.
Segment lengths provide a gauge of how aggressive the segmenter is.
A reasonable length for segments is 10-20 seconds.
EOS latency is measured as the delay from the end-of-speech (obtained via forced alignment) and the EOS timestamp.
For consistency, we only keep the EOS latencies for the last segment of each utterance in order to de-confound the fact that different segmenters have different number of segments.
1st pass WER is important for live-streaming accurate results to the user as they speak,
while 2nd pass WER is important for sending accurate results to the downstream task.

\begin{table}
\caption{Segment lengths (SL), EOS latencies (EOS), and WERs of various segmenters on YT\_LONG}
\vspace{-8pt}
\label{tab:main}
\centering
\resizebox{.99\columnwidth}{!}{%
\begin{tabular}{l|cc|cc|cc}
\toprule
                        &      &       &              &              & 2nd pass       & 1st pass           \\
Segmenter               & SL50 & SL90  & EOS50        & EOS90        &          WER   &          WER       \\
\midrule                
B3: Fixed-3s            & 3.0  & 3.0   & -            & -            & 19.60          & 19.20              \\
B4: Fixed-5s            & 5.0  & 5.0   & -            & -            & 17.78          & 19.39              \\
B5: Fixed-10s           & 10.0 & 10.0  & -            & -            & 16.82          & 19.16              \\
\midrule                
B6: VAD                 & 3.3  & 14.0  & 380          & 490          & 16.23          & 19.13              \\
E3: E2E                 & 6.9  & 20.8  & \textbf{240} & \textbf{480} & \textbf{15.86} & \textbf{18.72}     \\
$\Rightarrow$\;\; E3 vs. B6 & - & - & \textcolor{red}{-140} & \textcolor{red}{-10} & \textcolor{red}{-2.4\%} & \textcolor{red}{-2.1\%}  \\
\bottomrule
\end{tabular}%
}
\vspace{-6pt}
\end{table}

\begin{figure}
  \centering
  \vspace{-6pt}
  \includegraphics[width=0.45\columnwidth]{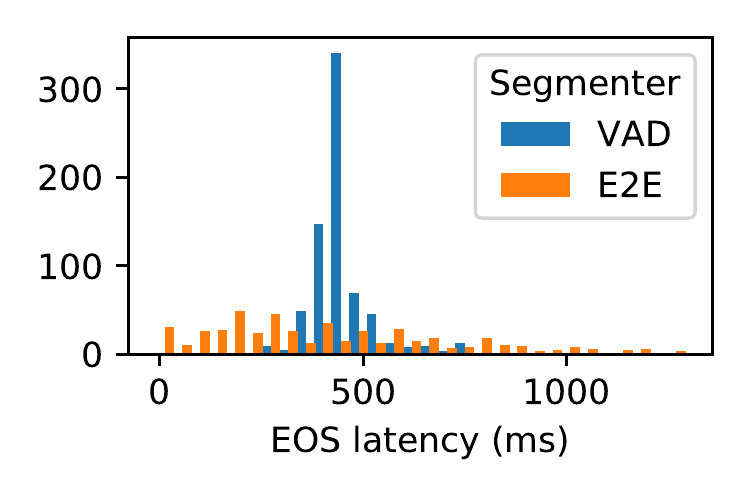}
  \includegraphics[width=0.45\columnwidth]{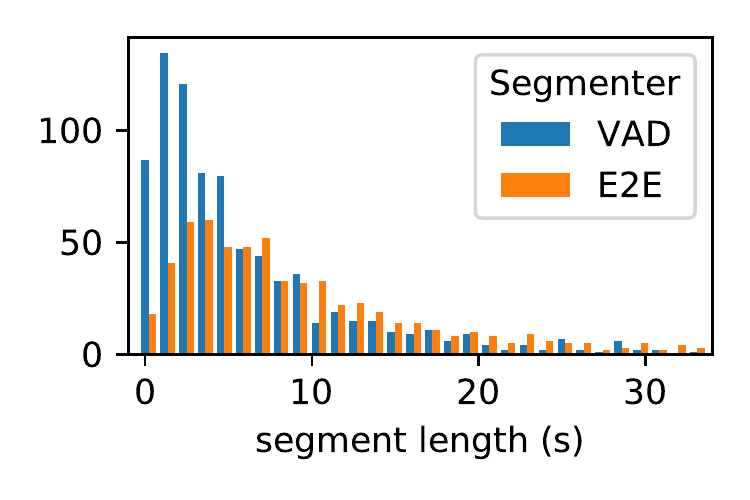}
  \vspace{-11pt}
  \caption{Distributions of latencies and segment lengths.}
  \label{fig:distro}
  \vspace{-14pt}
\end{figure}

\vspace{-5pt}
\subsection{Main results}
\vspace{-1pt}
Table \ref{tab:main} shows our main results: we evaluate the E2E segmenter (E3) against the VAD (B6) and a few fixed-length segmenters (B3-B5) as baselines.
Note that henceforth we will use E2E and VAD to refer to their respective segmenters.
The fixed-length segmenters fare worse than the VAD and E2E in WER as expected, despite having segment lengths within the same range,
indicating that segment length alone does not have a strong effect on WER.
The E2E segmenter achieves a median EOS latency gain of 140 ms, and a small 90th percentile gain of 10 ms over the VAD.
Further, it achieves a WER gains of 2.4\% and 2.1\% on the 2nd and 1st pass, respectively.
While the WER gains are modest, the median latency gain of 140 ms is significant for user experience.


The distributions of EOS latency are shown in Figure \ref{fig:distro} (top) for E2E and VAD.
Note here we include the EOS latencies across all segments in the testset (as opposed to only the last segment) to get more statistics.
E2E has significantly more variance in its EOS latency distribution,
perhaps reflecting the diversity of acousto-linguistic situations under which it must make a decision.
For example, some EOS may be easy to predict from the linguistic context,
while others may be ambiguous and require more acoustic context.
Note that the VAD has no EOS latency under 200 ms as expected due to its design of requiring 200 ms of detected silence before activating (Section \ref{sec:intro}).
The distributions of segment length are shown in Figure \ref{fig:distro} (bottom).
The biggest difference is that E2E has much fewer segments lengths near zero than the VAD,
reflecting the fact that during very long silences,
the VAD---which only cares whether the past 200 ms is silence---can emit more than one EOS signal within the same span of silence.

\subsection{Ablation study}
\begin{figure}
  \vspace{-5pt}
  \centering
  \includegraphics[width=.41\columnwidth]{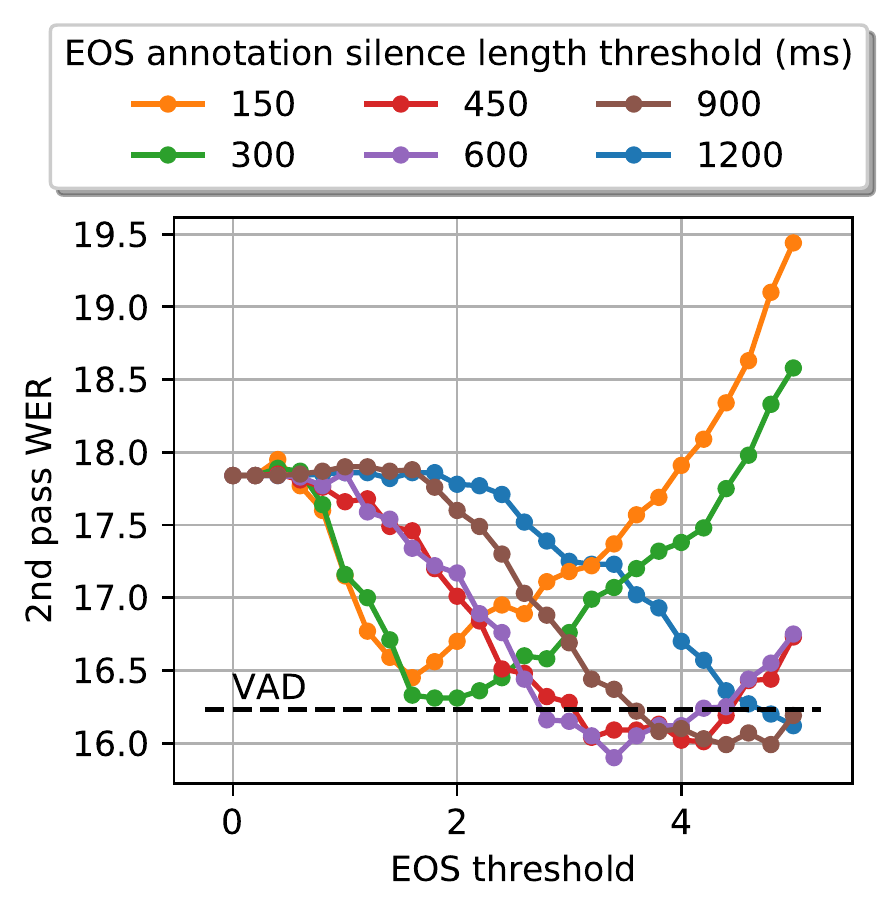}
  \quad
  \includegraphics[width=.41\columnwidth]{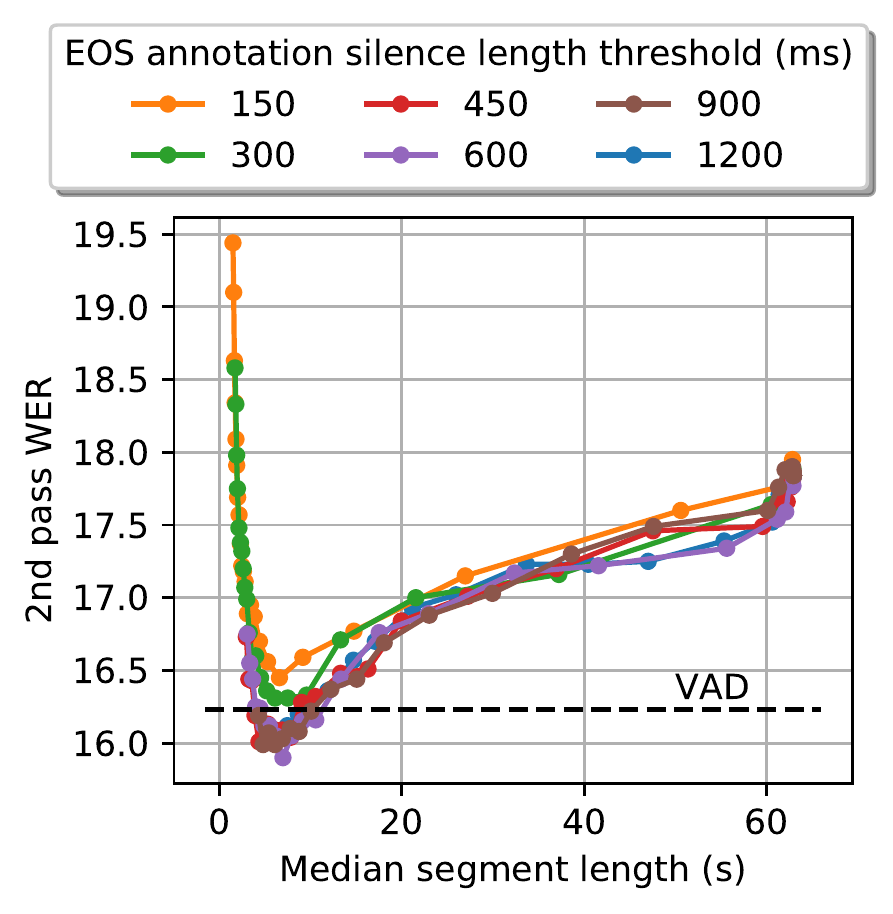}
  \vspace{-8pt}
  \caption{Ablation study of 2nd pass WER as a function of various levels of EOS emission aggressiveness during training and inference.
  }
  \label{fig:aggressiveness}
  \vspace{-12pt}
\end{figure}

There are two hyperparameters that tune how aggressively EOS signals are emitted during inference:
(1) \textbf{EOS threshold}, the negative-log-probability threshold under which EOS is emitted during \textit{inference}, and
(2) \textbf{Silence length threshold}, the minimum length of silence between two words in a \textit{training} utterance
such that a EOS token will be injected between those two words in its target transcript.
We provide an ablation study against both these parameters in Figure \ref{fig:aggressiveness} (left), where we plot the WER against the EOS threshold (x-axis) for multiple silence length thresholds (legend).
Each curve has a sweet spot where the WER is optimal,
and the EOS threshold at which this occurs gets smaller (tighter) as the silence length threshold gets smaller (looser).
It makes sense: if the model is trained to emit EOS more aggressively (smaller silence length threshold), then it needs to be more restricted during inference (smaller EOS threshold).
This study shows the silence length threshold hyperparameter is rather insensitive during training as we can compensate for it during inference.
In Figure \ref{fig:aggressiveness} (right), we plot the same data but against the median segment length (x-axis), which is tuned via the EOS threshold.
This shows that the changing the silence length threshold does not affect the optimal segment lengths during inference,
suggesting that the testset has an underlying sweet spot for how long segments should be.
We choose as our operating point the point that gives best WER, 600 ms silence length threshold and 3.7 EOS threshold,
and use this operating point for the results in Table \ref{tab:main} and Figure \ref{fig:distro}.

\vspace{-5pt}
\subsection{Oracle WER}
Since we use a fixed ASR model, measuring the oracle WER (OWER) can give some insight into how much improvement can come from a better beam search decoding.
In Table \ref{tab:oracle}, we report WER and OWER for baseline (B7-B9) and E2E segmenter (E4) on a 15-utterance subset of YT\_LONG.
This was the subset on which we were able to successfully obtain forced alignments on the transcripts, which are required for computing the oracle WER on long utterances without hitting memory limits.
In general, more frequent segmentation increases diversity in the beam search, but can also lead to poorer hypotheses.
To rule out this confounding factor, we do a controlled experiment where we pick a EOS threshold such that the median segment length is the same as that of the VAD.
At this setting, the E2E achieves a 5.6\% OWER gain over the VAD (Standard column in Table \ref{tab:oracle}).
Next, we repeat the comparison under a high lattice diversity scenario by applying 2-gram-based path merging during beam search \cite{prabhavalkar2021less} (Path-merging column).
We find that the OWER improves by 25\% relative to standard beam search due to path merging,
and under these conditions, the E2E achieves an even bigger OWER gain of 7.3\% over the VAD.
These findings suggest that segmentation improves ASR quality \textit{not} by creating more lattice diversity, but by improving the overall accuracy of the hypotheses in the lattice.

\begin{table}
\caption{Oracle WER from a standard and 2-gram-based path-merging beam search for various segmenters}
\vspace{-8pt}
\label{tab:oracle}
\centering
\resizebox{.78\columnwidth}{!}{%
\begin{tabular}{l|c|cc|cc}
\toprule
                        &      & \multicolumn{2}{c}{Standard}         & \multicolumn{2}{c}{Path-merging}   \\
Segmenter               & SL50 &          WER &         OWER          &          WER &         OWER        \\
\midrule
B7: Fixed-3s            & 3.0  & 11.6         & 10.3                  & 11.4         & 7.2                 \\
B8: Fixed-5s            & 5.0  & 9.6          & 7.8                   & 9.4          & 6.0                 \\
\midrule
B9: VAD                 & 4.4  & 7.8          & 5.4                   & 7.9          & 4.1                 \\
E4: E2E                 & 4.4  & 7.8          & \textbf{5.1}          & 7.9          & \textbf{3.8}        \\
$\Rightarrow$\;\; E4 vs. B9 & - & - & \textcolor{red}{-5.6\%} & - & \textcolor{red}{-7.3\%}   \\
\bottomrule
\end{tabular}%
}
\vspace{-10pt}
\end{table}

\vspace{-3pt}
\subsection{Error analysis}
Lastly, we investigate the decoding of some single utterances with the E2E and VAD segmenters to better understand where the wins and losses are coming from.
Table \ref{tab:error} shows snippets of three long utterance decodings for the E2E and VAD.
EOS locations are shown in blue and deletions/insertions relative to ground truth are shown in red/green.
In Table \ref{tab:error}a, the VAD substitutes ``up'' for ``just'', but E2E does not make that mistake.
The only difference was that the VAD emitted an EOS immediately after ``up'', causing the prior (erroneous) hypothesis to be finalized,
whereas E2E had the opportunity to correct the top hypothesis given later context.
A similar story applies to Table \ref{tab:error}b, where the VAD prematurely finalizes the deletion of ``you can put the'' by emitting an EOS immediately afterward.
The examples from Table \ref{tab:error}a and \ref{tab:error}b highlight a more general trend that the WER wins from E2E tend to be due to smarter EOS boundaries, preventing finalization of the wrong hypotheses.
We were unable to find any trends that favored the VAD over E2E,
and the VAD wins tended to be less systematic.
For example, Table \ref{tab:error}c shows a loss for E2E, where ``over stretch'' is substituted by ``overstretch''.
Since both VAD and E2E shared the same ASR model and differ only in the segmenters,
this error can only be attributed to the two having different decoder states and beam search lattices caused by earlier differences in EOS emission.
It's unlikely that such non-systematic errors favor one segmenter over the other.

\newcommand{\eos}{\textcolor{blue}{\textless{}EOS\textgreater}}
\newcommand{\insertion}[1]{\textcolor{green}{#1}}
\newcommand{\deletion}[1]{\textcolor{red}{\st{#1}}}

\begin{table}[b]
\small
\vspace{-14pt}
\caption{Error analysis.}
\vspace{-8pt}
\label{tab:error}
\centering
\resizebox{.99\linewidth}{!}{%
\begin{tabular}{l|l|l}
\toprule
          & VAD Segmenter                                                                       & E2E Segmenter                                            \\
\midrule                      
(a) Win  & \ldots\eos the inch \eos                                                            & \ldots\eos the inch \eos                                   \\
         & increments here all my line                                                         & increments here all my line                                \\
         & \eos \deletion{just} \insertion{up} \eos                                            & \eos just                                                  \\
         & so we get the same \eos \ldots                                                      & so we get the same \eos \ldots                             \\
\midrule                       
(b) Win  & \ldots you can go around like i was                                                 & \ldots you can go around like i was                        \\
         & doing before \deletion{you} \deletion{can} \deletion{put}                           & doing before \eos you can put                              \\
         & \deletion{the} \eos edge every time                                                 & \deletion{the} \insertion{today} edge every time           \\
         & right up to your \ldots                                                             & right up to your \ldots                                    \\
\midrule
(c) Loss & \ldots\eos you don't have to go                                                     & \ldots\eos you don't have to go                            \\
         & back and forth and                                                                  & back and forth and                                         \\
         & over stretch                                                                        & \deletion{over} \deletion{stretch} \insertion{overstretch} \\
         & or i'm sorry over shrink or                                                         & or i'm sorry over shrink or                                \\
         & over stretch                                                                        & \deletion{over} \deletion{stretch} \insertion{overstretch} \\
         & depending what you're doing and                                                     & depending what you're doing and                            \\
         & cause an issue \eos \ldots                                                          & cause an issue \eos \ldots                                 \\
\bottomrule
\end{tabular}
}
\end{table}
\vspace{-5pt}
\section{Conclusion}
We have unified the E2E segmenter with the two-pass cascaded encoder model,
an important step toward low latency and high quality long-form speech recognition in streaming mode.
Advantages of this unification include simplicity (removing additional VAD module), EOS latency reduction, and WER improvements on both the 1st and 2nd pass.

\vfill\pagebreak

\bibliographystyle{IEEEbib}
\bibliography{refs}

\end{document}